\documentclass{article}
\usepackage{spconf,amsmath,graphicx,CJKutf8,tabularx}

\title{DECOUPLING RECOGNITION AND TRANSCRIPTION IN MANDARIN ASR}
\name{Jiahong Yuan$^1$, Xingyu Cai$^1$, Dongji Gao$^2$, Renjie Zheng$^1$, Liang Huang$^1$, Kenneth Church$^1$}
\address{$^1$Baidu Research USA\\
$^2$Johns Hopkins University}
\begin{document}
%\ninept
%
\maketitle
\begin{abstract}
Much of the recent literature on automatic speech recognition (ASR) is taking
an end-to-end approach. Unlike English where the writing system is closely related to sound, Chinese characters (Hanzi) represent meaning, not sound. We propose factoring audio $\rightarrow$ Hanzi into two sub-tasks: (1) audio $\rightarrow$ Pinyin and (2) Pinyin $\rightarrow$ Hanzi, where Pinyin is a system of phonetic transcription of standard Chinese. Factoring the audio $\rightarrow$ Hanzi task in this way achieves 3.9\% CER (character error rate) on the Aishell-1 corpus, the best result reported on this dataset so far.

% Automatic speech recognition (ASR) maps from speech audio to orthographic transcription. From a human’s point of view, the task consists of two cascade processes. From speech to phonological representation, which is recognition, and then from phonological representation to orthographic transcription. In this paper, we proposed a framework to decouple recognition and transcription in Mandarin ASR. We demonstrated that the framework could significantly improve Mandarin ASR compared to an end-to-end approach. Our system achieved 3.9\% CER (character error rate) on the Aisell-1 corpus, which is the best reported result on the dataset to date. 
\end{abstract}
\section{Introduction}
\label{sec:intro}

Automatic speech recognition (ASR) maps from speech audio to orthographic transcription. For languages with a phonemic writing system such as Spanish and English, the mapping is largely from an acoustic stream to a phonological representation. For languages with a logographic writing system such as Mandarin Chinese, however, the mapping is more complicated due to the arbitrary association between sounds and Chinese characters. 

In conventional ASR systems, a pronunciation lexicon or model is used to bridge orthographic words and phonemes or other phonological representations. The difference in the type of orthography only requires building language-specific pronunciation lexicons or models, but the framework of separating acoustic, pronunciation, and language models is equally valid for languages with different writing systems.

In recent years, end-to-end approaches to automatic speech recognition have gained popularity and achieved great success. End-to-end speech recognition does not use a lexicon. It learns a mapping between speech audio and orthographic transcription entirely and directly from paired data. For English, letters or sub-words have been used as targets to learn by a network model. These targets are largely phonological units (although English does not have a one-to-one correspondence between sounds and letters). For Mandarin, the targets are naturally Chinese characters. Chinese characters are not phonological units. From our analysis of a large corpus, a tonal syllable in Mandarin Chinese corresponds to 7 Chinese characters on average, and the number may be as high as 80 for some syllables (see results in Table~\ref{tab:staticstics} below). Apparently, the burden is much higher for a network to learn a mapping between audio and Chinese characters, compared to English letters.

In this study, we promote the use of phonological units, represented by Pinyin, as targets in Mandarin speech recognition. We propose a framework for Mandarin ASR that consists of two cascade components: from audio to Pinyin, which we call “recognition”, and from Pinyin to Chinese characters, called “transcription”. We argue that the recognition component is comparable to end-to-end English ASR. The transcription component is independent and should be decoupled from recognition. Our method achieved 3.9\% CER (character error rate) on the Aisell-1 corpus, the best reported result on the dataset so far. 

\section{Related work}
\label{sec:work}

\subsection{State of the art of Mandarin ASR}

Transformer-based models have become popular recently. The numbers below summarize 30 arXiv papers \cite{arxiv1,arxiv2,arxiv3,arxiv4,arxiv5,arxiv6,arxiv7,arxiv8,arxiv9,arxiv10,arxiv11,arxiv12,arxiv13,arxiv14,arxiv15,arxiv16,arxiv17,arxiv18,arxiv19,arxiv20,arxiv21,arxiv22,arxiv23,arxiv24,arxiv25,arxiv26,arxiv27,arxiv28,arxiv29,arxiv30} in the last 12 months that reported performance on Aishell 1, a benchmark dataset for Mandarin ASR.  There were 106 index terms in these 30 papers.  After “speech recognition”, the most common index terms are:

\begin{itemize}

\item \textbf{transformer (17):} transformer (12), speech transformer (2), transformer-xl (1), spike triggered non-autoregressive transformer (1), convolution-augmented transformer (1)

\item \textbf{end-to-end (14):} end-to-end speech recognition (5), end-to-end (4), end-to-end asr (1), robust end-to-end speech recognition (1), end-to-end speech processing (1), end-to-end models (1), streaming end-to-end (1)

\item \textbf{attention (12):} self-attention (2), attention (1), self-attention mechanism (1), self-attention network (1), gaussian-based self-attentin (1), multi-head attention (1), hybrid ctc and attention (1), ctc-attention (1), online attention (1), stochastic attention head removal (1), MoChA (1)

\item \textbf{non-autoregressive (8):}  non-autoregressive (7), spike triggered non-autoregressive transformer (1)

\item \textbf{ctc (6):} connectionist temporal classification (2), ctc (1), ctc module (1), ctc-attention (1), hybrid ctc and attention (1)

\item \textbf{conformer (4):} conformer (3), convolution-augmented transformer (1)

\item \textbf{streaming (4):} streaming speech recognition (3), streaming end-to-end (1)

\end{itemize}

From the list, we can see that Transformer-based models are predominately used. 20 of the 28 papers adopted Transformer \cite{vaswani2017attention} or Conformer \cite{Gulati2020ConformerCT}, i.e., convolution-augmented Transformer. Transformer is a sequence-to-sequence model that has three components: encoder, attention, and decoder. The encoder extracts acoustic features and the decoder extracts language features and predicts output sequence, both with the help of self-attention. The alignment between acoustic and language features are learned by encoder-decoder attention. From the point of view of conventional ASR, the three components of Transformer function as acoustic modeling, language modeling, and pronunciation modeling, respectively. Therefore, the model can be, and has been, naturally applied for end-to-end speech recognition.

Much effort has been made to improve the architecture of Transformer, originally proposed for machine translation, to make it more suitable for speech recognition. Conformer \cite{Gulati2020ConformerCT}, for example, combines Transformer with CNN (convolutional neural network) to model both global and local dependencies of an audio sequence. \cite{arxiv8} conducted experiments on 17 ASR corpora, including Aishell-1, to compare the performance of Transformer and Conformer models. Their results showed that Conformer outperformed Transformer on 14 of the 17 corpora. On Aishell-1 test set, the character error rate was 6.7\% for Transformer and 4.7\% for Conformer.

CTC (Connectionist Temporal Classification) \cite{graves2006connectionist} is another frequent keyword on the list. Like attention, CTC can do alignment between input and output that have different lengths (such as audio and linguistic labels), and has been widely used in end-to-end speech recognition. With CTC, fine-tuning pre-trained wav2vec2.0 \cite{baevski2020wav2vec} models has achieved state-of-the-art performance for ASR in English and low resource languages \cite{Yi2020ApplyingWT}. We have found that the method also performed well on emotion recognition \cite{Cai2021Emotion} and recognition of suprasegmental units such as syllables, tones, and pitch accents \cite{Yuan2021Suprasegmentals}. In this paper, we adopted Wav2vec2.0 with CTC for “recognition” and Transformer for “transcription” in our proposed framework.

Most of the surveyed papers (21 of 30) reported a character error rate between 5\% and 7\% on Aishell-1 test set \cite{Bu2017AISHELL1AO}, an improve over the baseline system published with the dataset in Kaldi \cite{Povey_ASRU2011} (i.e., error rate of 7.9\% using Chain TDNN model).  Five papers reported error rates under 5\% \cite{arxiv8,arxiv16,arxiv24,arxiv25,arxiv30}. The best character error rate of 4.5\% was reported in \cite{arxiv24}, which combines conformer with WFST (weighted finite-state transducers) based non-autoregressive decoding. \cite{arxiv30} proposed a method to correct errors in ASR output. They reported character error rate of 4.8\% using the Conformer architecture and 4.1\% after error corrections. 

\subsection{Modeling units in Mandarin ASR}

In conventional, hmm-based ASR systems, modeling units are phonological representations. For Mandarin ASR, phonological units such as phonemes, syllable initials/finals, and syllables have been employed \cite{Chen1997NewMI,Hwang2009MadarinASR}. A central issue in choosing modeling units for Mandarin ASR is how to model tones \cite{Chang2000LargeVM,Lei2006ToneASR}. A number of approaches have been proposed in the literature, by either modeling tones independently from phones or using tone-dependent phonological units such as tonal finals or tonal syllables. 

End-to-end ASR directly models graphemes or other units in orthography. For English, letters, sub-words, and words have been used. \cite{Li2019BytesAA} proposed to use Unicode bytes as a representation of text. Their results showed that bytes are superior to grapheme characters over a wide variety of languages in monolingual end-to- end speech recognition. Characters are the most commonly used modeling unit for end-to-end ASR in Mandarin Chinese; sub-words have also been employed \cite{Zhang2019TowardsLM}.

\cite{Li2015ACS} compared the performance of syllables, initials/finals, and phones in deep neural networks (DNNs) based Mandarin ASR. Their results showed that context dependent phones obtained the best performance. \cite{Zhou2018ACO} studied five modeling units in Transformer-based Mandarin ASR: context-independent phonemes, syllables, words, sub-words and characters. They showed that character-based model performed best. \cite{Chan2016OnOA} reported that joint character-pinyin training improved character error rate over the character only model in an attention-based model for online Mandarin speech recognition. \cite{Zhang2019InvestigationOM} proposed hybrid Character-Syllable modeling units by mixing high frequency Chinese characters and syllables, and demonstrated that using these units in CTC based acoustic modeling for Mandarin speech recognition could dramatically reduce character error rate. \cite{Fu2020ResearchOM} compared the performance of different modeling units in RNNT (Recurrent Neural Network Transducer) based Mandarin ASR. They found that using syllables with tone outperformed the use of initials and finals with tone with an average of 13.5\% relative character error rate reduction. In this study, we tried both syllables with and without tone as modeling units for recognition in our proposed framework.

\section{Proposed framework}
\label{sec:framework}

The proposed framework is illustrated in Figure 1. 

\begin{figure}[htb]
  \centering
  \includegraphics[width=1.0\linewidth]{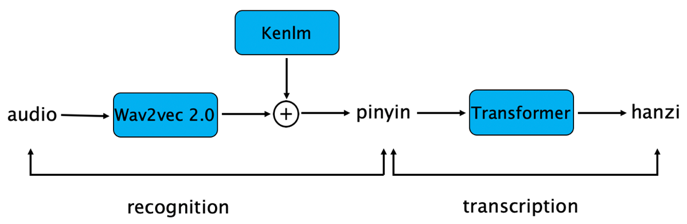}
  \caption{The framework of Mandarin ASR with decoupled recognition and transcription components.}
  \vspace{-0.1in}
  \label{fine-tuning}
\end{figure}

The recognition module employs Wav2vec2.0 with CTC, fine-tuned using paired audio and pinyin data, for recognizing pinyin from audio. A pinyin language model, i.e., a language model trained on sentences in pinyin, may also be employed in decoding. The following transcription module converts the pinyin output into Chinese characters (i.e., Hanzi) through a Transformer model. 

To train a pinyin language model (KenLM n-gram) and Pinyin-Hanzi Transformer we only need text data. Because large-scale text corpora are easily available, these models can be trained on a much larger dataset than is used for fine-tuning Wav2vec2.0.

Several similar frameworks have been proposed in the literature. \cite{Zhang2020DecouplingPA} proposed a decoupled transformer model to use monolingual paired data and unpaired text data to alleviate the problem of data shortage in end-to-end ASR system for code-switching. The model consists of two parts: audio-to-phoneme (A2P) network and phoneme-to-text (P2T) network, and they are optimized jointly through attention fusion. \cite{Pham2020IndependentLM} proposed to separate the decoder subnet from the encoder output in end-to-end speech recognition. In their framework, the decoupled subnet is an independently trainable LM subnet, which can easily be updated using extra text data. \cite{Wang2021CascadeRS} proposed to use an RNN-T to transform acoustic feature into syllable sequence, and then convert the syllable sequence into character sequence through an RNN-T-based syllable-to-character converter.

\section{Experiments and results}
\label{sec:experiments}

\subsection{Datasets}

Our experiments were conducted on Aishell-1, a benchmark dataset for Mandarin ASR. Aishell-1 contains 165 hours of read speech in Mandarin Chinese from 400 speakers. The speakers are from different dialect regions but most are from northern areas. The corpus includes training (150 hours), development (10 hours), and test (5 hours) sets.

Two additional text corpora, the Lancaster Corpus of Mandarin Chinese and Chinese Gigaword Fifth Edition, were used to train language models and Pinyin-to-Hanzi transformer models. We selected sentences between 5 and 40 characters from these corpora, and made sure that the sentences did not appear in the development or test set of Aishell-1. 

\subsection{Fine-tuning Wav2vec 2.0}

The procedure of fine-tuning Wav2vec2.0 is illustrated in Figure 2. A randomly initialized linear projection is added on top of the contextual representations of Wav2vec 2.0 to map the representations into modeling units, and the entire model is optimized by minimizing the CTC loss through fine-tuning. 

\begin{figure}[htb]
  \centering
  \includegraphics[width=1.0\linewidth]{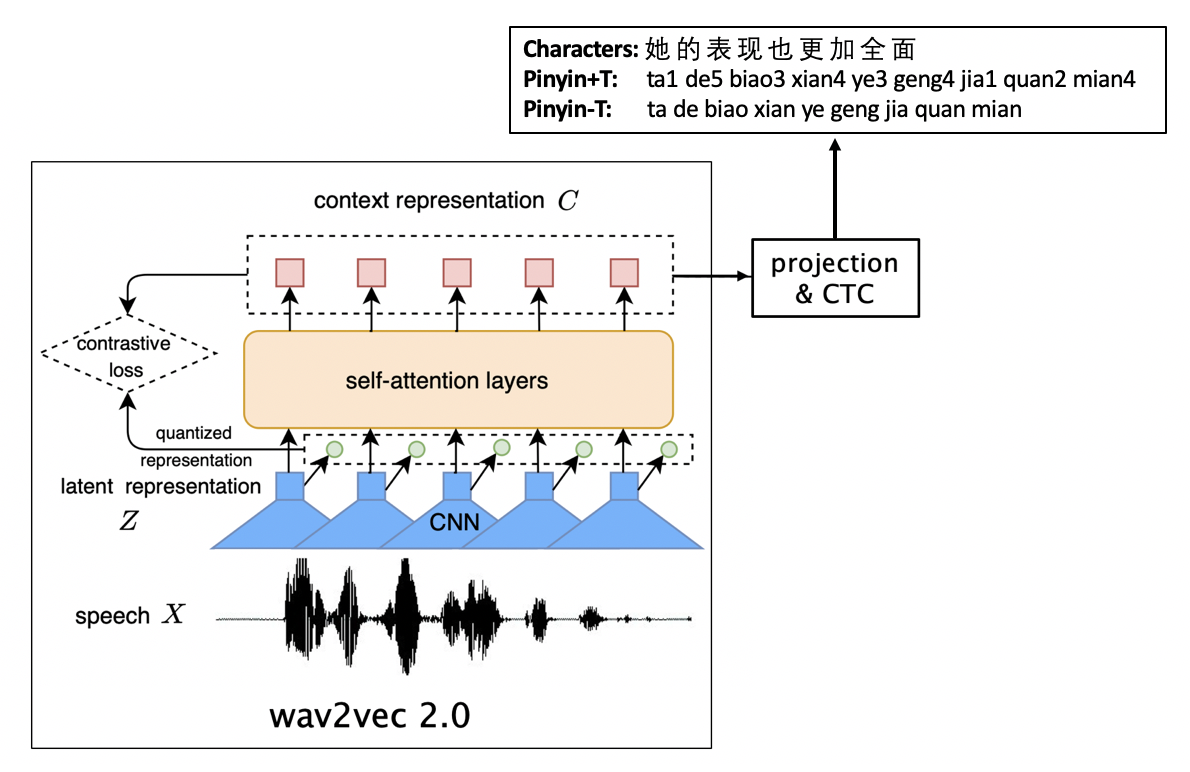}
  \caption{The framework of fine-tuning Wav2vec 2.0 for pinyin or character recognition.}
  \vspace{-0.1in}
  \label{fine-tuning}
\end{figure}

We tried both pinyin with tones (Pinyin+T) and without tones (Pinyin-T) as modeling units for Pinyin recognition. To compare with the proposed framework of decoupling recognition and transcription, we also tried Chinese characters as modeling units, using Wav2vec2.0 to directly recognize Chinese characters. The number of modeling units of characters, pinyin syllables with and without tone is 4333, 2020, and 408, respectively.

To be used for decoding Wav2vec2.0 with CTC, KenLM 6-gram language models were trained for different modeling units, respectively. For pinyin language models, Chinese characters were converted into Pinyin syllables using a python package called \textit{pypinyin}. Every character or pinyin is an independent unit and word boundaries are not present in the training data. A dummy lexicon, in which every modeling unit has a pronunciation of itself, is used for decoding with the language model. 

The experiments were conducted using \textit{fairseq}. In all experiments, the wav2vec 2.0 large model pre-trained on 960 hours of Librispeech audio, libri960\_big.pt, was used for fine-tuning. For the first 10k updates only the output classifier is trained, after which the Transformer is also updated. The max\_tokens was set to 1.1 million (which is equivalent to 68.75-second audio with sampling rate of 16 kHz), the learning rate was 5e-5. The development set was used to determine the number of updates, which is 300k (~109 epochs) for Pinyin+T, 470K (~170 epochs) for Pinyin-T and 460k (~167 epochs) for characters.

Table~\ref{tab:fine-tuning} lists the results, from using no language model and language models trained on progressively increasing amounts of text data. When no language model was used, Pinyin+T had a unit error rate 4.3\%. If we remove tone marks from the output of Pinyin+T, the units become Pinyin-T with an error rate of 2.6\%, lower than that from using Pinyin-T as modeling units. The same result can also be seen for language models trained on different amounts of data. For example, the Pinyin-T unit error rate was 1.7\% from the Pinyin+T model and 2.0\% from the Pinyin-T model, when the language models were trained on 90M sentences. This result suggests that there is an interaction between tones and segments, and therefore segments can be better distinguished from each other with tone. 

When no language model was used, the unit error rate was 4.3\%, 2.8\%, and 7.2\% for Pinyin+T, Pinyin-T, and characters, respectively. Language models helped the recognition of these units. With a language model trained on 90 million sentences, the Pinyin+T, Pinyin-T, and Character error rate dropped to 2.5\%, 2.0\%, and 5.3\%, respectively. 

\begin{table} [t]
\caption{Results of fine-tuning Wav2vec2.0 for recognition of different units. The numbers in parentheses are Pinyin-T errors, calculated by removing tone marks from Pinyin+T output.}
\label{tab:fine-tuning}
\begin{center}
\begin{tabular}{|c|c|c|c|c|}
\hline
 \multicolumn{2}{|c|}{Training data} & \multicolumn{3}{|c|}{Unit error rate} \\
Wav2vec2.0 & KenLM & Pinyin+T & Pinyin-T & Char. \\
\hline
& No LM & 4.3\% & 2.8\% & 7.2\% \\
&  & (2.6\%) & &  \\
\cline{2-5}
& 120k & 3.6\% & 2.7\% & 6.8\% \\
& Aishell-1 & (2.3\%) &  &  \\
\cline{2-5}
Aishell-1 & 1M & 3.0\% & 2.3\% & 5.9\% \\
 &  & (1.9\%) &  &  \\
\cline{2-5}
& 10M & 2.8\% & 2.2\% & 5.7\% \\
&  & (1.8\%) & &  \\
\cline{2-5}
& 90M & 2.5\% & 2.0\% & 5.3\% \\
&  & (1.7\%) &  &  \\
\hline
\end{tabular}
\end{center}
\end{table}

\subsection{Pinyin to Hanzi}

The pinyin output from the recognition module needs to be converted into Chinese characters. This task can be seen as a translation from pinyin to Chinese characters. To understand the complexity of this task, we did an analysis of the mapping from pinyin and Chinese characters, by counting the number of possible outputs in Chinese characters given a pinyin sequence. The results, based on 90 million sentences, are listed in Table~\ref{tab:staticstics}, for both pinyin with and without tone.

\begin{table} [t]
\caption{Statistics of mapping from pinyin N-gram, with or without tone, to Chinese characters. The numbers in parentheses are for pinyin without tone.}
\label{tab:staticstics}
\begin{center}
\begin{tabular}{|c|c|c|c|c|}
\hline
N-gram  & Average & Maximum & Percentage \\
\hline
1-gram	& 7.68	& 82 & 16.9\% \\
        & (23.83)	& (156)	& (3.7\%) \\
\hline
2-gram	& 4.71	& 302	& 30.4\% \\
        & (28.19)	& (1380) & (6.0\%) \\
\hline
3-gram	& 1.50	& 177	& 74.1\% \\
        & (3.64) & (870) & (43.6\%) \\
\hline
4-gram	& 1.11	& 43	& 92.1\% \\
        & (1.39) & (182) & (79.5\%) \\
\hline
5-gram	& 1.03	& 28	& 97.4\% \\
        & (1.09) & (63)	& (93.3\%) \\
\hline
6-gram	& 1.01	& 14	& 99.0\% \\
        & (1.03) & (27)	& (97.6\%) \\
\hline
\end{tabular}
\end{center}
\end{table}

From Table~\ref{tab:staticstics} we can see that a 1-gram pinyin with tone corresponds to 7 Chinese characters on average, and the number may be as high as 80 for some 1-grams. Only 16.9\% of 1-grams correspond to only one character. A 6-gram pinyin with tone corresponds to 1.01 sequences of characters on average, and 99.0\% of 6-grams correspond to only one sequence of characters. Without tone, a 1-gram pinyin corresponds to 23.83 Chinese characters on average, and 97.6\% of 6-grams correspond to only one sequence of characters. 

We trained Transformer models to convert pinyin to Chinese characters, for pinyin with and without tone, respectively. We used the same architecture and hyperparameters as \textit{transformer\_iwslt\_de\_en} in \textit{fairseq}. Both models were trained on paired pinyin and characters of 90 million sentences for 20 epochs. Table~\ref{tab:pinyin2hanzi} lists the results of testing on a held-aside dataset which consists of 14k sentences.

\begin{table} [t]
\caption{Performance of Transformer models converting pinyin to Chinese characters.}
\label{tab:pinyin2hanzi}
\begin{center}
\begin{tabular}{|c|c|}
\hline
Input units	& output character error rate \\
\hline
Pinyin+T	& 1.2\% \\
Pinyin-T	& 3.6\% \\
\hline
\end{tabular}
\end{center}
\end{table}

\subsection{Evaluation}

We evaluated the performance of the proposed framework and compared it to fine-tuning wav2vec2.0 to directly output Chinese characters, an end-to-end approach. The results of using pinyin+T and pinyin-T as recognition units are listed in Table~\ref{tab:evaluation+T} and Table~\ref{tab:evaluation-T}, respectively. From Table~\ref{tab:evaluation+T}, we can see that using pinyin+T, our proposed framework significantly outperformed the baseline end-to-end approach. With a KenLM language model trained on additional text data of 90 million sentences, our method achieved 3.9\% character error rate on Aishell-1 test set, a 35\% error reduction over the end-to-end approach. It is also the best reported result on the benchmark dataset so far.

When pinyin-T is used as recognition units, however, the performance of the proposed framework underperformed the baseline end-to-end approach, as shown in Table~\ref{tab:evaluation-T}.  

\begin{table} [t]
\caption{Character error rate on Aishell-1 test set, using pinyin+T as recognition units in the proposed framework.}
\label{tab:evaluation+T}
\begin{center}
\begin{tabular}{|c|c|c|c|}
\hline
\multicolumn{2}{|c|}{Training data} & \multicolumn{2}{|c|}{Character error rate} \\
Wav2vec2.0	& KenLM	& End-to-end & Pinyin+T \\
\hline
&	No LM	& 7.2\%	& 6.1\% \\
\cline{2-4}
Aishell-1 & Aishe1l-1 &	6.8\% & 5.3\% \\
\cline{2-4}
 & 90M	& 5.3\% &  \textbf{3.9\%} \\
\hline
\end{tabular}
\end{center}
\end{table}

\begin{table} [t]
\caption{Character error rate on Aishell-1 test set, using pinyin-T as recognition units in the proposed framework.}
\label{tab:evaluation-T}
\begin{center}
\begin{tabular}{|c|c|c|c|}
\hline
\multicolumn{2}{|c|}{Training data} & \multicolumn{2}{|c|}{Character error rate} \\
Wav2vec2.0	& KenLM	& End-to-end & Pinyin-T \\
\hline
&	No LM	& 7.2\%	& 7.3\% \\
\cline{2-4}
Aishell-1 & Aishe1l-1 &	6.8\% & 7.1\% \\
\cline{2-4}
 & 90M	& 5.3\% &  6.0\% \\
\hline
\end{tabular}
\end{center}
\end{table}

\section{Conclusions}
\label{sec:conclusions}
Mandarin ASR maps from audio to Hanzi. In this study, we proposed a framework that decomposes the task into two sub-tasks:
audio $\rightarrow$ Pinyin ("recognition) and Pinyin $\rightarrow$ Hanzi ("transcription"). Our method achieved 3.9\% character error rate on the Aishell-1 corpus, the best reported result on the dataset so far.

We compared the performance of using pinyin with (pinyin+T) and without (pinyin-T) tone as recognition units in our proposed framework. Although the recognition-unit error rate was slightly lower when pinyin-T was employed, converting from pinyin-T into Chinese characters generated more errors. As a result, using pinyin+T outperformed the baseline end-to-end approach with a large margin (i.e., 35\% relative error reduction) whereas using pinyin-T underperformed the baseline. From the application point of view, new models need to be developed to improve the accuracy of converting pinyin-T into Chinese characters. From a linguistic point of view, however, pinyin-T is an incomplete representation of sound in Mandarin Chinese and therefore should not be used as recognition units in our proposed framework.

\bibliographystyle{IEEE}
\bibliography{mybib}

\end{document}